\documentclass[letterpaper, 10 pt, conference]{ieeeconf} 

\IEEEoverridecommandlockouts 
\makeatletter

\let\proof\@undefined
\let\endproof\@undefined
\makeatother

\usepackage{mathptmx} 
\usepackage{amssymb} 
\usepackage{amsthm}
\usepackage{amsmath} 

\usepackage{booktabs}
\usepackage{verbatim}

\usepackage{xcolor}
\usepackage{subfig}
\usepackage{graphicx}
\usepackage{caption}
\usepackage{array}

\usepackage{siunitx}

\usepackage{hyperref} 
\hypersetup{
    colorlinks,
    citecolor=black,
    filecolor=black,
    linkcolor=black,
    urlcolor=black,
    pdfauthor={},
    pdfsubject={},
    pdftitle={}
}

\usepackage{algorithm, algorithmic}
\usepackage{array}
\usepackage{url}

\usepackage{cleveref}
\newcommand*\phantomrel[1]{\mathrel{\phantom{#1}}}

\def\figSoftHex{\includegraphics[width=1.0\columnwidth]{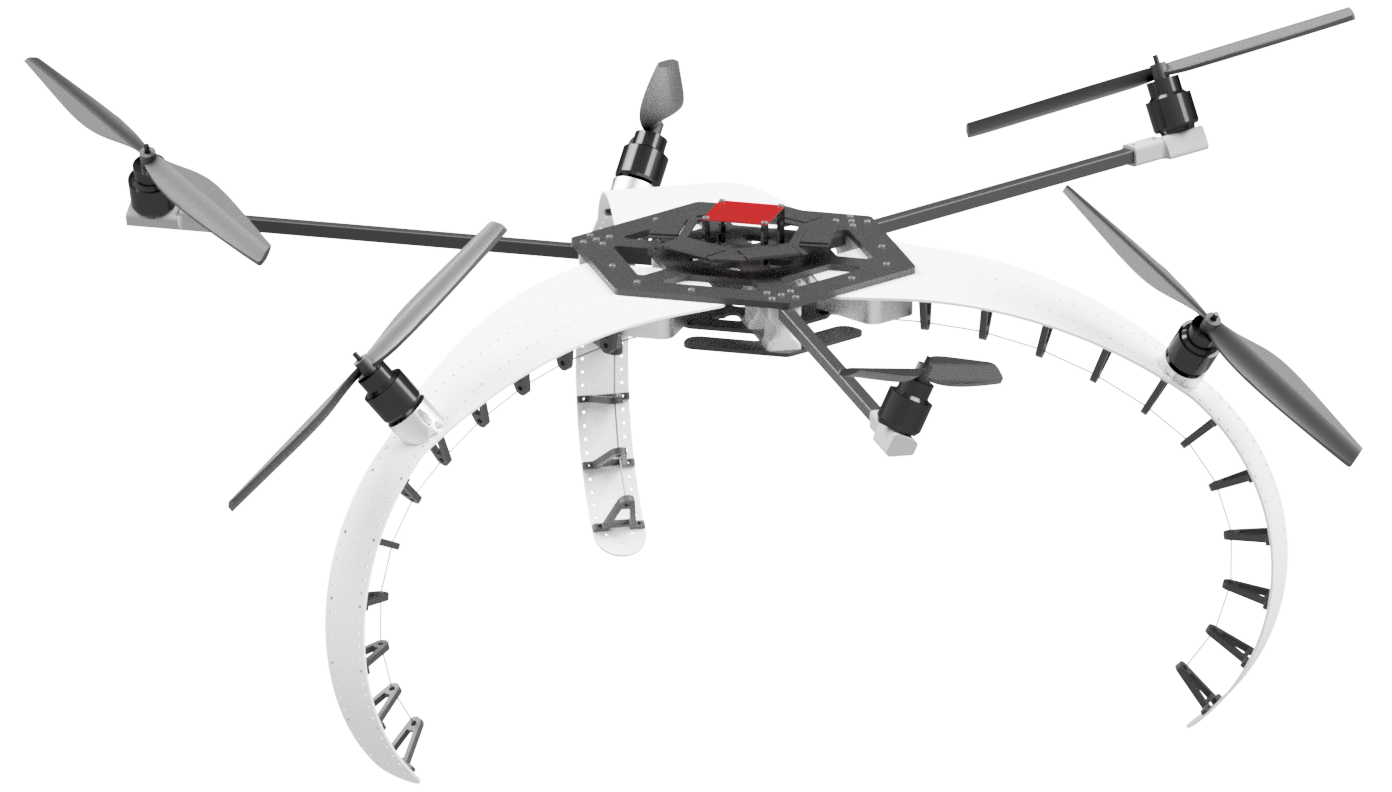}}
\def\figSchematics{\includegraphics[width=1.0\columnwidth]{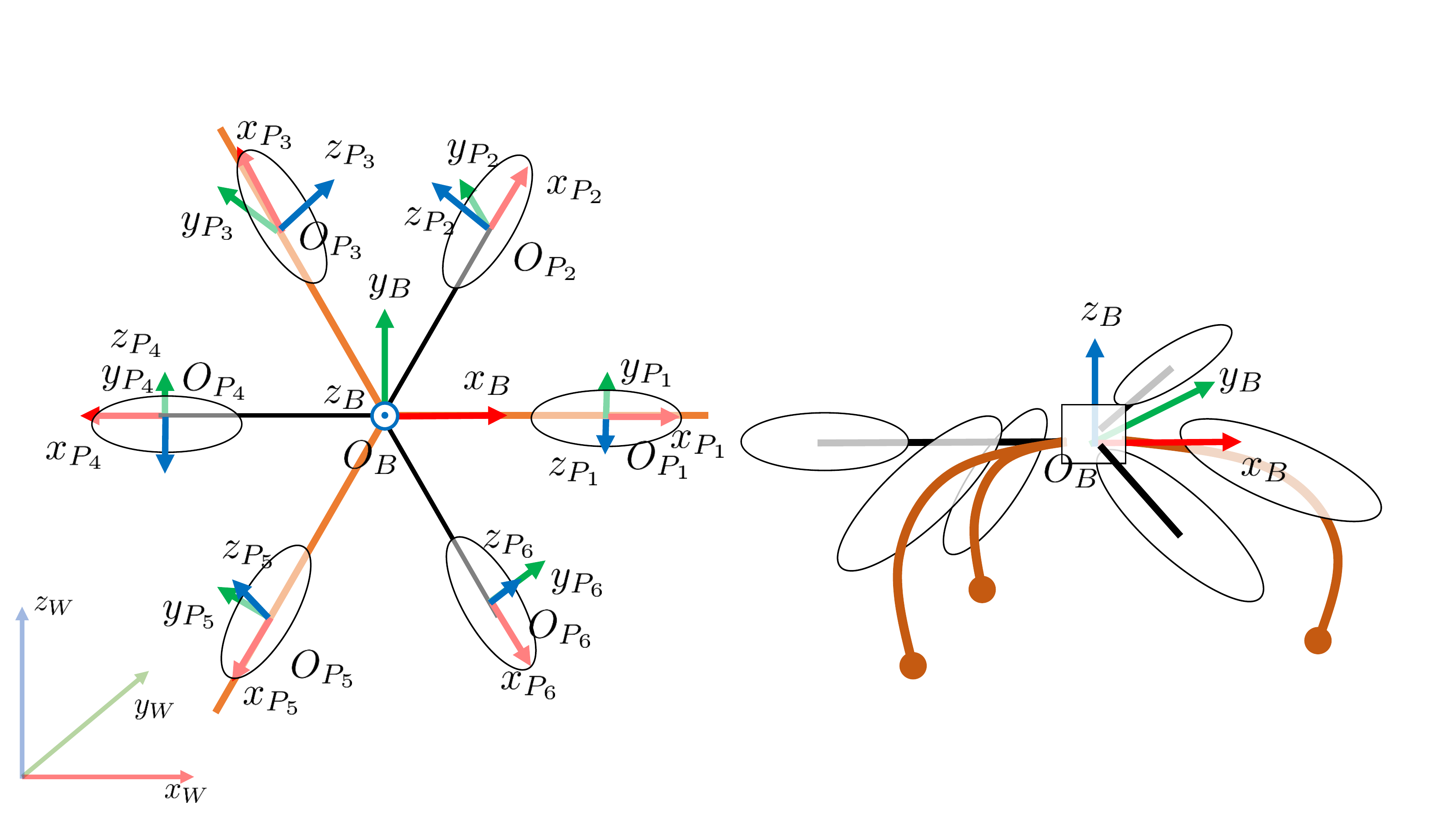}}
\def\figSoftArm{\includegraphics[width=1.0\columnwidth]{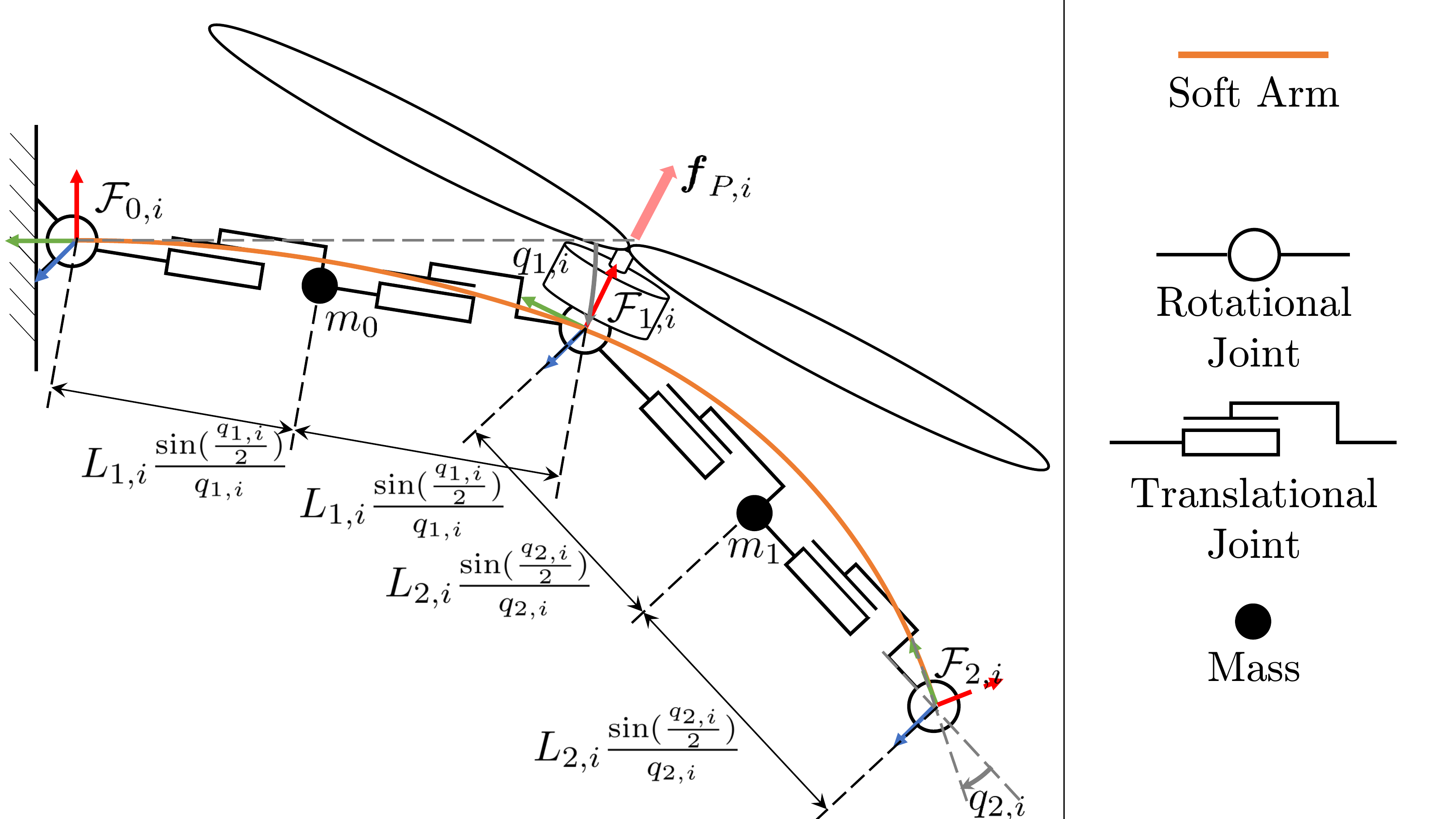}}

\def\figExpOnePos{\includegraphics[width=0.92\columnwidth]{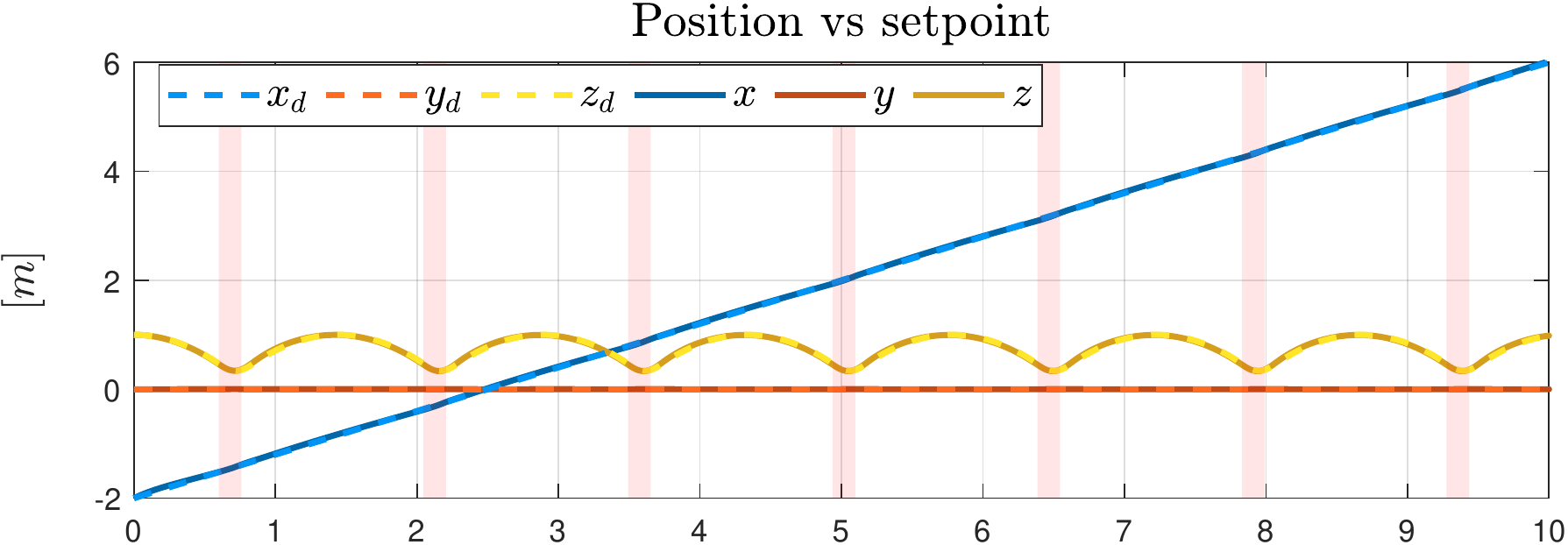}}
\def\figExpOneAtt{\includegraphics[width=0.92\columnwidth]{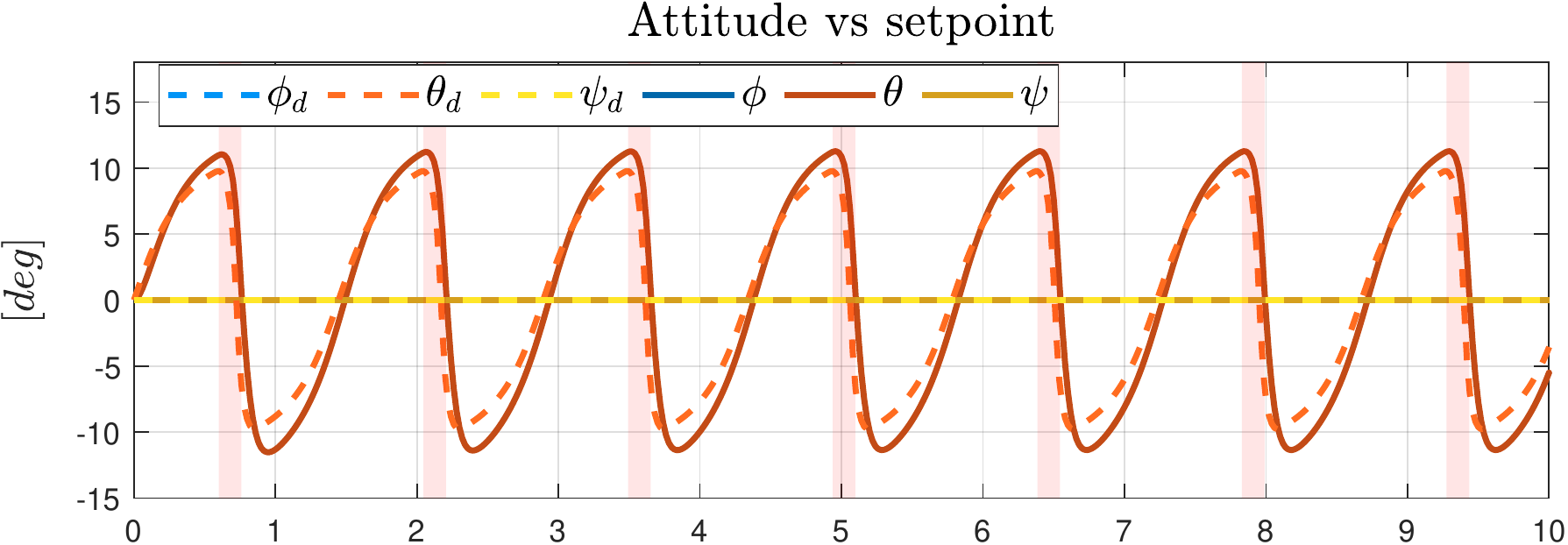}}
\def\figExpOneeP{\includegraphics[width=0.92\columnwidth]{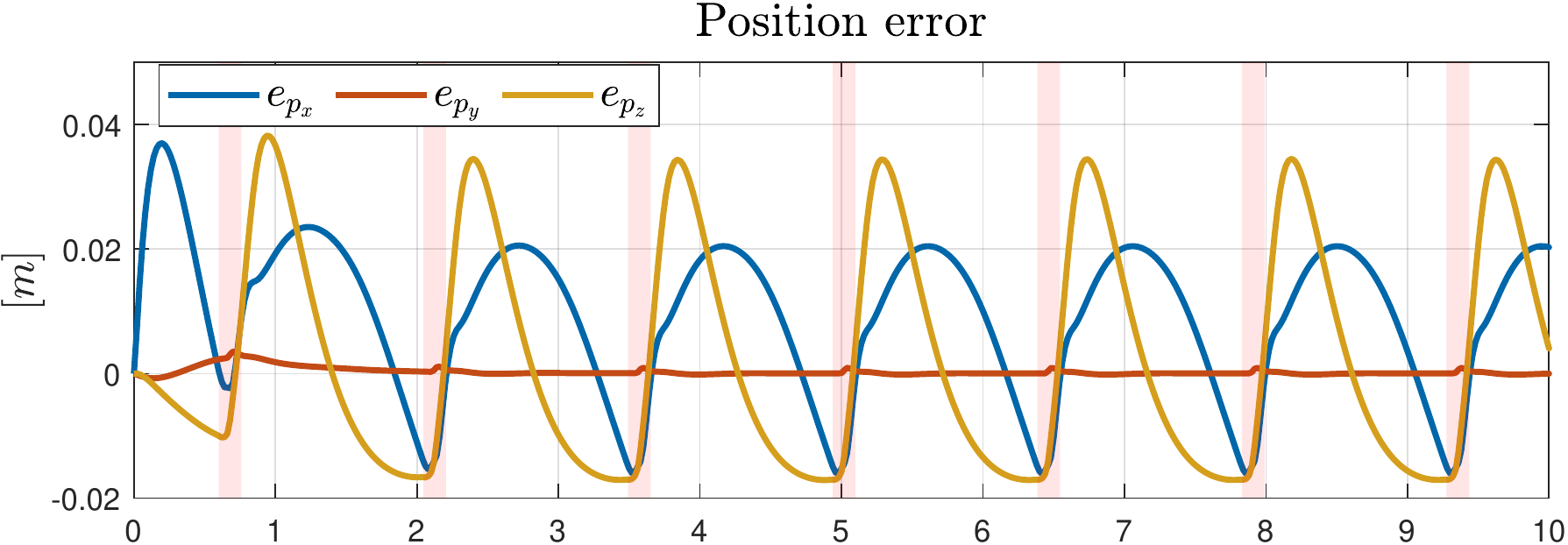}}
\def\figExpOneeR{\includegraphics[width=0.92\columnwidth]{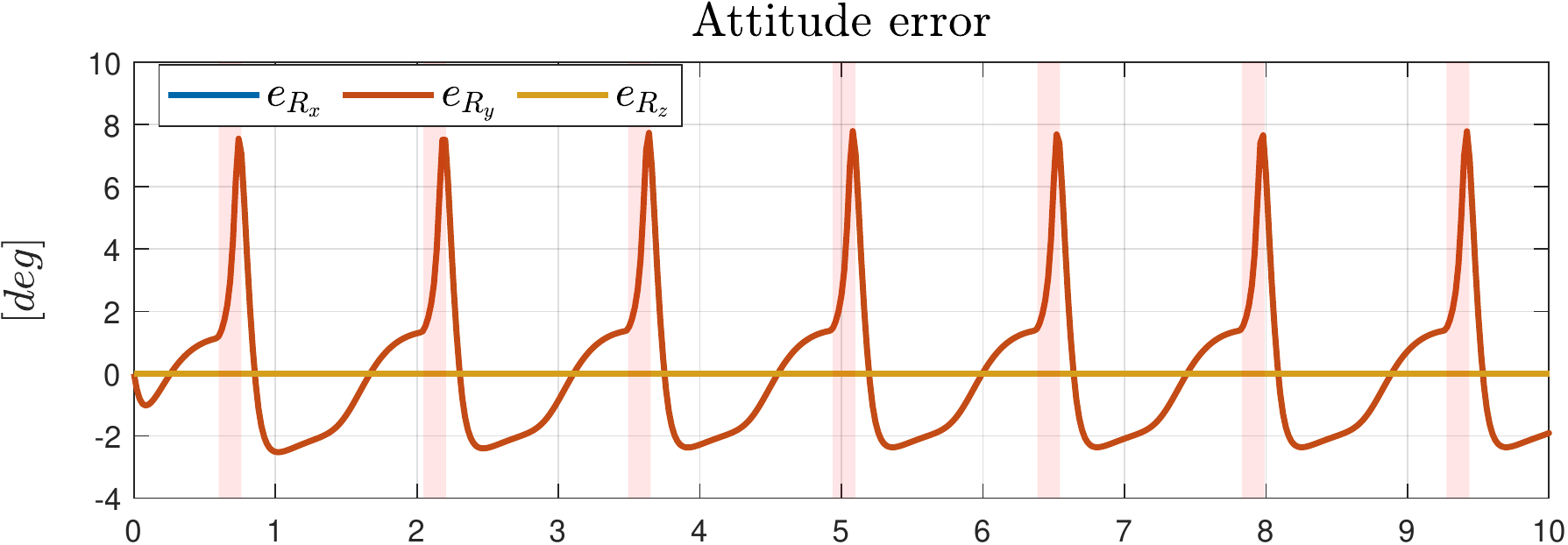}}
\def\figExpOneArms{\includegraphics[width=0.92\columnwidth]{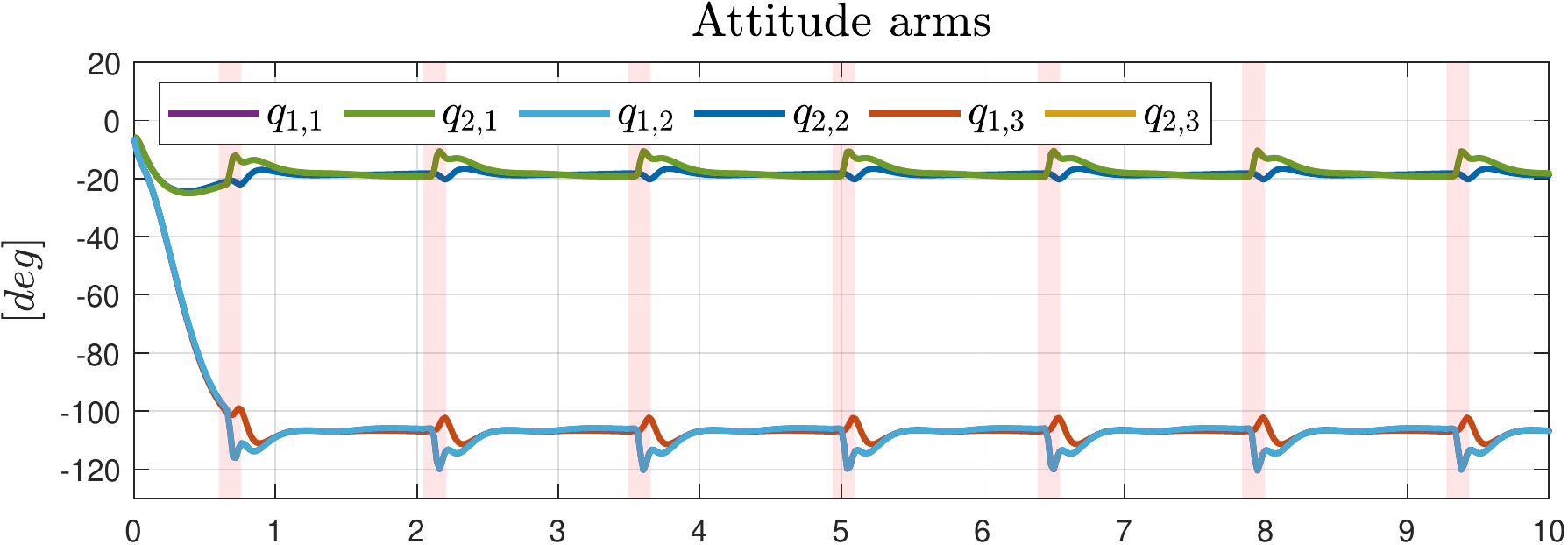}}
\def\figExpOneProps{\includegraphics[width=0.92\columnwidth]{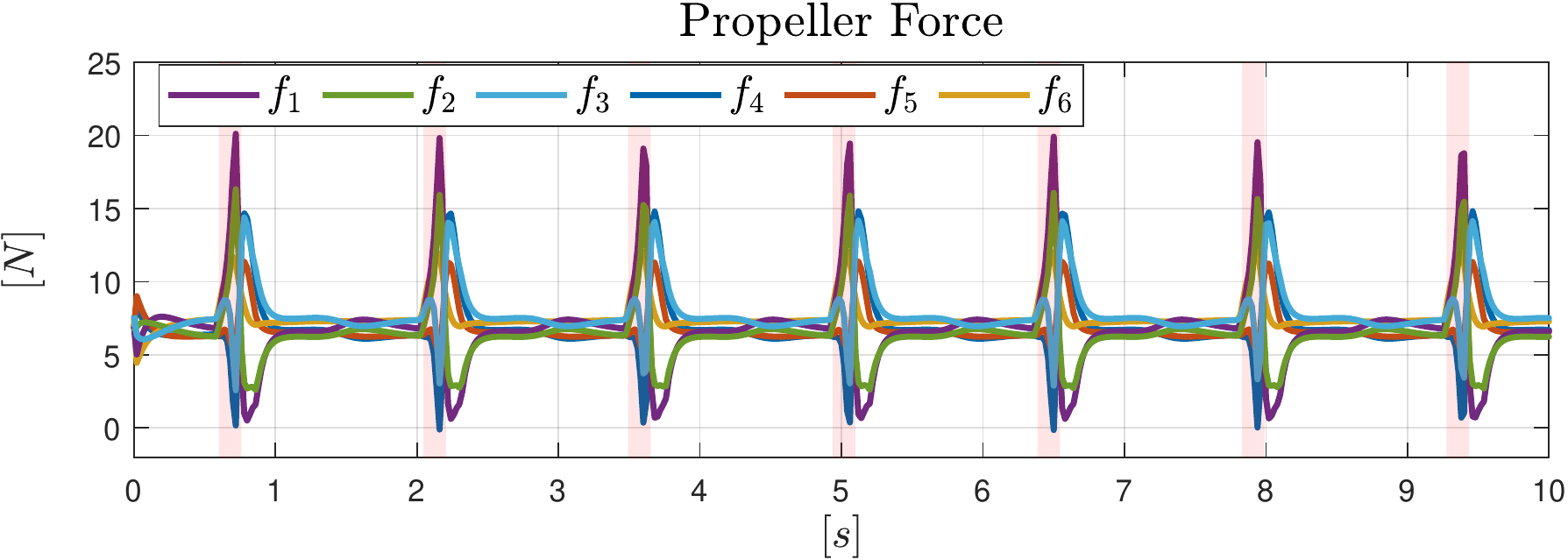}}

\def\figEnergy{\includegraphics[width=1.0\columnwidth]{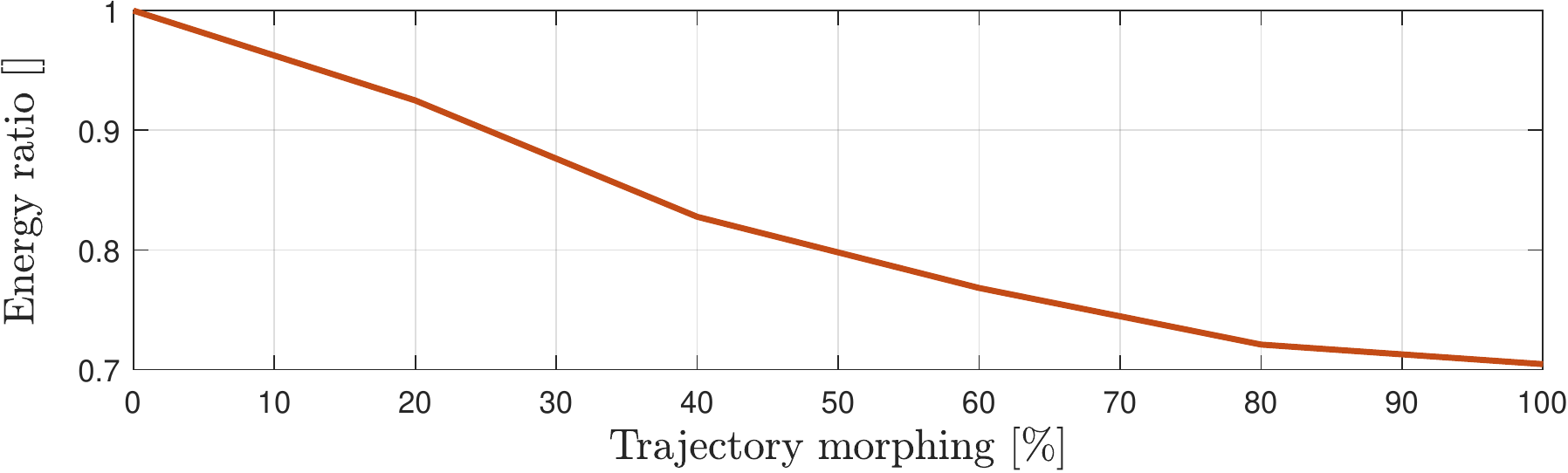}}

\makeatletter
\renewcommand{\dddot}[1]{%
  {\mathop{\kern\z@#1}\limits^{\vbox to-1.4\ex@{\kern-\tw@\ex@
   \hbox{\normalfont ...}\vss}}}}
\renewcommand{\ddddot}[1]{%
  {\mathop{\kern\z@#1}\limits^{\vbox to-1.4\ex@{\kern-\tw@\ex@
   \hbox{\normalfont....}\vss}}}}
\makeatother

\newcommand{\vect}[1]{\mathbf{#1}}
\newcommand{\matr}[1]{\mathbf{#1}}

\newcommand{\nR}[1]{\mathbb{R}^{#1}}		

\newcommand{\norm}[1]{\left\lVert#1\right\rVert}	

\newcommand{\wFrame}{\mathcal{F}_W}		
\newcommand{\bFrame}{\mathcal{F}_B}		

\newcommand{\vx}{\vect{x}}
\newcommand{\vy}{\vect{y}}

\newcommand{\vz}{\vect{z}}

\newcommand{\vomegaB}{\boldsymbol{\omega}_B}	

\newcommand{\rotMatB}{\matr{R}_B}

\title{\LARGE \bf{SMORS: A soft multirotor UAV for multimodal locomotion and robust interaction}} 

\author{Markus Ryll$^1$, and Robert K. Katzschmann$^2$
\thanks{$^1$Autonomous Aerial Systems, Department of Aerospace and Geodesy, TU München, Willy-Messerschmitt-Str. 1,  
82024 Taufkirchen/Ottobrunn, Germany, {\tt \footnotesize \href{mailto:markus.ryll@tum.de}{markus.ryll@tum.de}}\newline
$^2$Soft Robotics Lab, Department of Mechanical and Process Engineering, ETH Zurich, Tannenstrasse 3, 8092 Zurich, Switzerland, {\tt \footnotesize \href{mailto:rkk@ethz.ch}{rkk@ethz.ch}}}
}

\makeatletter
\def\ps@titlepagestyle{
	\def\@oddhead{\textcolor{red}{\sf\footnotesize Preprint version, final version accepted for ICRA 2022 \hfill}}
	\def\@evenhead{\textcolor{black}{\sf\footnotesize Preprint version, final version accepted for ICRA 2022 \hfill}}
}%
\makeatother
\makeatletter
\def\ps@headings{
	\def\@oddfoot{\textcolor{black}{\sf\footnotesize Preprint version, final version accepted for ICRA 2022 \hfill}}
	\def\@evenfoot{\textcolor{black}{\sf\footnotesize Preprint version, final version accepted for ICRA 2022 \hfill}}
}%
\makeatother
\begin{document}
\maketitle 

\begin{abstract}
We present \textit{SMORS}, the first \textit{S}oft fully actuated \textit{M}ultir\textit{O}to\textit{R}  \textit{S}ystem for multimodal locomotion. 
Unlike conventional hexarotors, SMORS is equipped with three rigid and three continuously soft arms, with each arm hosting a propeller.
We create a bridge between the fields of soft and aerial robotics by mechanically coupling the actuation of a fully actuated flying platform with the actuation of a soft robotic manipulator.
Each rotor is slightly tilted, allowing for full actuation of the platform. 
The soft components combined with the platform's full actuation allow for a robust interaction, in the form of efficient multimodal locomotion.
In this work, we present the dynamical model of the platform, derive a closed-loop control, and present simulation results fortifying the robustness of the platform under a jumping-flying maneuver. We demonstrate in simulations that our multimodal locomotion approach can be more energy-efficient than the flight with a hexarotor. 
\end{abstract}
\vspace{-2mm}

\section{Introduction}
In opposition to rigid robots, most natural organisms are soft and compliant and perform their tasks precisely. Soft organisms solve dynamic tasks efficiently thanks to their built-in elasticity. Inspired by nature, researchers developed a broad spectrum of soft or compliant robots\,\cite{Rus2015, Rich2018}. 
Current multirotor platforms, \emph{e.g.}, quadrotors, are rigid robots. Aerial multirotor platforms have become popular and widely used in research and industry over the last decade. While initially mainly interesting for contact-free monitoring like visual inspection\,\cite{6496959} or search and rescue tasks\,\cite{tian2020search}, aerial multirotor platforms have recently been used for manipulation\,\cite{8299552, ryll20196d} and transportation\,\cite{thiels2015use} tasks. Underactuated multirotor platforms have been equipped with manipulators to enable grasping tasks\,\cite{6631278}. In the next iteration step, fully actuated and overactuated aerial platforms have been developed\, \cite{ryll2014novel, rashad2020fully, brunner2020trajectory}. These aerial platforms can track an arbitrary wrench profile (independent force and torque profile), allowing for a decoupled tracking of a position and orientation trajectory. This tracking capability allows for a broader spectrum of tasks, as fully actuated platforms tend to be able to track trajectories more precisely, allow for direct rejection of disturbances\,\cite{rajappa2015modeling, jiang2014nonparallel}, and enable the application of an arbitrary wrench on the environment\,\cite{8435987}.

\begin{figure}[t]
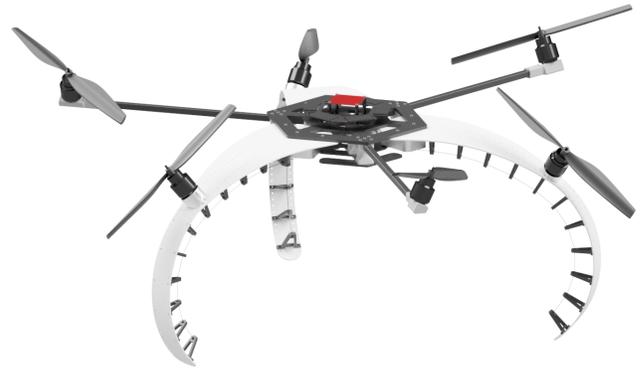

\centering
\figSoftHex
\caption{Computer-aided design (CAD) model of SMORS. The fully-actuated hexarotor has three rigid (black) and three compliant arms (white). The compliant arms are actuated by the aerodynamic forces of the propeller plus an additional actuator, allowing for deflection over the full arm length. Tilted propellers render the platform to what is known as fully actuated in the field of multirotor systems.}
\label{fig:SoftHex}
\vspace{-5mm}
\end{figure}

In this work, we aim at bridging the world between soft and aerial robotics by combining a fully actuated platform with soft robotic manipulators to enable multimodal locomotion and robust object grasping. Therefore, we present SMORS, a \textbf{S}oft fully actuated \textbf{M}ultir\textbf{O}to\textbf{R} \textbf{S}ystem with three soft continuous manipulators. To our knowledge, this is the first work combining soft continuous manipulators with aerodynamic actuators, namely the propellers. Previous work in the field have only equipped multi-rotor platforms with compliant actuators, but have not fully integrated the two into a unified concept. Towards adaptive compliance in environmental interaction, Hamaza et al. equipped a UAV with a passive, compliant manipulator\,\cite{hamaza2018towards}. In \cite{7139280, 7353585}, the authors designed an aerial system with an active compliant actuator for safe physical interaction. All systems discussed so far offer joint compliance but do not have continuously deformable manipulators. 
Mishra et al. demonstrated in \cite{mishra2018design} an image-based visual servo scheme for a hexarotor equipped with a soft grasper to perform object detection and grasping.
Fishman et al. demonstrated in \cite{fishman2020control, fishman2021dynamic} grasping with a standard underactuated quadrotor and four soft grippers. The work proposed an approach to the control and trajectory planning of quadrotors enabling object grasping.

With this work, we break a new frontier in this line of research by presenting the first partially soft multirotor platform. We enable (i) more robust and, therefore, safer interactions and (ii) present a multimodal locomotion approach for a soft multirotor. This paper specifically contributes:
\begin{itemize}
\item A description and modeling of the first aerial robot with soft continuous actuators hosting its propellers.
\item A closed-loop 3D pose controller for tasks such as jumping and object grasping.
\item The validation of the controller in a simulated jumping motion. We can show experimentally, that the platform is more energy-efficient than conventional flying.
\end{itemize}

The paper structure continues as follows. First, we describe the kinematic and dynamic model of the system. Next, the controller is introduced and simulated experimental results are presented. Finally, an outlook on on-going work is given.

\section{Problem Statement and Design Decisions of the Soft Multirotor}
For safe interactions with their environments, rigid robots require exact information of their state and the state of their surroundings. Already small errors can cause an instability of the system and catastrophic oscillations. 
A robotic platform should ideally be capable of selectively softening and stiffening its externally facing components to guarantee the fulfillment of the passivity principle.

We choose to merge the design concepts of a fully actuated hexarotor with a continuously deformable soft arm in a step towards this goal. In contrary to previous works, we do not heterogeneously attach a soft gripper below a rigid drone and deal with their modeling and control in a decoupled manner, but instead, we take three of the six thrust-generating propellers and attach those directly to three soft arms that are extruding radially from a UAV platform. The soft arm itself can be curled up or straightened out, allowing for a controlled variable tilt of the propellers. During a manipulation maneuver of our manipulator drone, the three fixed propellers provide the necessary lift, while the three propellers on the compliant arms allow for an additional grasping force to support the arms in bringing up enough force for a stable grasp.  

The curious reader might wonder what drove the design and kinematic decisions for the SMORS prototype. The design followed the following guidelines. (i) A fully actuated platform was required to enable full orientation and position tracking capabilities; (ii) we aim for static stability during contact with the ground, \emph{e.g.}, during landing and when grasping objects; and (iii) our goal is to achieve a natural damping of the system during locomotion by jumping.

To fulfill the above requirements, the following design decisions have been made. The fixed tilted hexarotor is from a mechanical and control perspective the simplest and most reliable aerial platforms for full actuation. Therefore this platform type was selected to fulfil guideline (i). At least three contact points are needed to meet guideline (ii). To reduce the total mass and complexity of the system, we selected the minimum number of actuated arms, i.e., three. We decided to couple the soft arms with the position of the motors to enable the desired dampening. During the jumping locomotion, the contact force aligns the propeller force vectors of the three soft arms automatically towards the rotor's z-direction. This allows for less required controller action as demonstrated in the experiments.

The soft arms can be physically implemented using either a tendon-driven, fluidic, electrostatic, or thermal actuator. Our initial design in \Cref{fig:SoftHex} relies on a lightweight and accessible tendon-driven design approach~\cite{odhner2014compliant}, due to comparably high force density, low payload, accessible electromagnetic drives, and easy integration with a UAV platform. Every soft arm consists of a flexible beam with an inlaid flexible constraint layer, allowing it to bend only around one axis. Reinforcing brackets are attached at equal separations along the flexible beam, providing structure and further limiting the bending motion to one axis. A tendon cable is routed along the flexible beam through openings within the brackets. At the proximal end of the flexible beam, the tendon cable is attached to a winch motor. When the winch of the motor is rolled up in one direction, the tendon cable applies a force via the tendon cable to the attachment point of the cable at the distal end of the flexible beam. This actuation causes the beam to curl downwards with constant curvature. When the winch is rotated in the other direction, the arm will fully straighten out.

\section{Modeling}
In this section, we discuss the kinematics and dynamics of the soft aerial vehicle. We will begin with the kinematics of the rigid main body and a single soft arm and then discuss the dynamics of the single components and finally derive the full system dynamics. For better understanding, we have numbered the six arms (three flexible, three rigid) with $i=1,\dots,6$, where odd numbers refer to soft arms and even numbers refer to rigid arms. 

\subsection{Kinematics}
\subsubsection{Rigid elements}
\begin{figure}[ht]
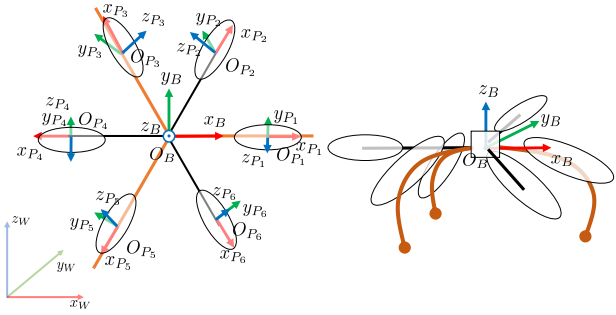

\centering
\figSchematics
\caption{Schematic view of the soft unmanned aerial vehicle. The rigid arms are colored in black and the soft arms are colored in orange. Left: Top view of the schematics with fully stretched arms; Right: Side view with actuated soft arms bending downwards.}
\label{fig:Schematics}
\end{figure}

Let us begin by defining the rigid parts of the system.
We consider a world frame $\wFrame=O_W,\{ \vx_W, \vy_W, \vz_W\}$ and place within the world frame a body frame $\bFrame=O_B,\{ \vx_B, \vy_B, \vz_B\}$ which defines the position and orientation of SMORS. The body frame is attached to the rigid body of the aerial platform, where $O_B$ corresponds to the geometric center of the three rigid propellers (see \Cref{fig:Schematics}). We further approximate that $O_B$ coincides with the center of mass (CoM) of the system. This is a fair assumption since computations in CAD show that the actual CoM is moved only marginally by motions of the arms.
$\vect{p}_B$ is the position vector from $O_W$ to $O_B$ expressed in $\wFrame$. The velocity of $O_B$ is expressed by $\vect{v}_B \in \nR{3}$, and the rotation matrix $\rotMatB \in \mathsf{SO}(3)$ defines the orientation of $\bFrame$ relative to $\wFrame$. The angular velocity of $\bFrame$ with respect to $\wFrame$, expressed in $\bFrame$, is denoted with $\vomegaB \in \nR{3}$. The kinematics of $\rotMatB$ are then
\begin{equation}
\dot{\vect{R}}_B = \vect{R}_B [{\boldsymbol\omega}_B]_{\times},
\end{equation}
where $[\bullet]_{\times}\in {\sf SO}(3)$ represents, in general, the skew symmetric matrix associated to any vector $\bullet \in \mathbb{R}^3$.

Let's generally define with $\vect{e}_1$, $\vect{e}_2$, and $\vect{e}_3$ the three vectors of the canonical basis of $\mathbb{R}^3$, and with $\vect{R}_x$, $\vect{R}_y$, and $\vect{R}_z$ the three canonical rotation matrices in ${\sf SO}(3)$.

Let us next define the three fixed propeller frames ${\cal F}_{P_2}, {\cal F}_{P_4}, {\cal F}_{P_6}$, where ${\cal F}_{P_i}=O_{P_i},\{ \vx_{P_i}, \vy_{P_i}, \vz_{P_i}\}$ with $i = 2,4,6$ coincides with the center of the rigidly mounted propellers. The orientations of ${\cal F}_{P_i}, i = 2, 4, 6$ are given by the rotations matrix 
\begin{align}
\vect{R}_{P_i}^B = \mathbf{R}_z\left((i-1)\frac{\pi}{3}\right) \mathbf{R}_{x}(\alpha),	\quad  i=2, 4, 6  \label{eq:RSi_def}
\end{align}
where $\mathbf{R}_x(\alpha)$ describes the fixed tilting of the propellers, guaranteeing full actuation of the platform at any feasible arm configuration.

The vector from $O_B$ to $O_{P_i}$, describing the position of the center of the $i$-th propeller, expressed in $\bFrame$, is 
\begin{align}
\vect{p}^B_{B,P_i}= 
\mathbf{R}_z\left((i-1)\frac{\pi}{3}\right) \vect{l}_i, \quad i=2,4,6
\end{align}
where  
$\vect{l} =[l\ 0\ 0]^T$  defines the length of the rigid arms. 

\subsubsection{Single soft arm}
\begin{figure}[ht]
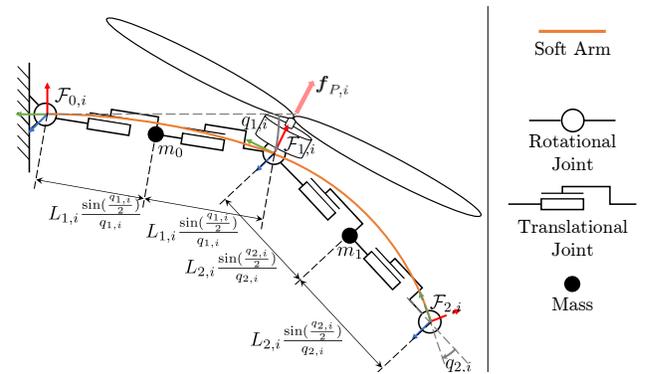

\centering
\figSoftArm
\caption{Kinematic motion of a single arm. Black dots indicate origin of arm, position of propeller and end-effector.}
\label{fig:SoftArm}
\end{figure}

Next, the kinematic properties of a single soft arm are discussed. We consider the arm to be continuously compliant about one axis while being rigid about the other two axes (see \Cref{fig:Schematics} and \Cref{fig:SoftArm}). The arm is actuated by one additional input, causing a deflection over the full length of the soft arm. Additionally, we consider the propeller to be mounted along the arm, causing an additional deflection between the root of the arm and the position of the propeller, due to the propeller's generated forces. With this setup, we can (i) actively change the tilting angle of the propeller, by moving the arm and (ii) actuate the arm by changing the thrust of the propeller.
To model the soft arm, the Piecewise Constant Curvature (PCC) model\,\cite{webster2010design} is utilized. Following the PCC approach, every single arm consists of two segments. The first segment, actuated by the additional input and the propeller forces, starts at the root of the arm and ends at the mounting point of the propeller. The second segment, only actuated by the additional input, starts at the propeller and ends at the end of the arm, namely the end-effector. Every segment has a constant curvature, hence the resulting curve is everywhere differentiable. We will now attach reference frames to the start- and end-points of every segment along a single arm. The position of the first frame $\mathcal{F}_{0, i}$, with $i=1,3,5$, is located at the root of the arm. $\mathcal{F}_{1, i}$ defines the mounting point of the propeller's motor, and frame $\mathcal{F}_{2,i}$ is defined by the end of the arm and therefore being the end-effector's contact point (see \Cref{fig:SoftArm}). Thanks to the constant curvature hypothesis, we use the relative rotation angle $q_{j,i}$ between the reference frames $\mathcal{F}_{j-1,i}$ and $\mathcal{F}_{j,i}$ for $j=1,2$ to fully define the arm's configuration. The set of two 2D homogeneous transformations, $\vect{T}^{0,i}_{1,i}$ and $\vect{T}^{1,i}_{2,i}$, describes the arm's kinematics as
\begin{equation}
\vect{T}^{j-1,i}_{j,i}=
\begin{bmatrix}
\cos(q_{j,i}) & -\sin(q_{j,i}) & L_{j,i}\frac{\sin(q_{j,i})}{q_{j,i}}\\
\sin(q_{j,i}) &   \cos(q_{j,i}) & L_{j,i}\frac{1-\cos(q_{j,i})}{q_{j,i}}\\
0 & 0 & 1
\end{bmatrix}
\label{eq:armKinematics}
\end{equation}
where $L_{j,i}$ is the nominal (non-deflected) length of the segment.

To enable full actuation of SMORS in any configuration of the arms, the propellers of the soft arms are also tilted by $\alpha$ but in opposite orientation compared to the fixed arms. The tilting direction is defined with respect to the tangent axis of the soft arm's bending (see\,\Cref{fig:SoftHex,fig:Schematics}). The orientation and position of the actuated propellers are therefore

\begin{align}
\vect{R}_{P_i}^B &=   \mathbf{R}_z\left((i-1)\frac{\pi}{3}\right) \mathbf{R}_y \Big(q_{j,i}\Big)\mathbf{R}_{x}(-\alpha),	\quad  i=1, 3, 5 \\
\vect{p}^B_{B,P_i}&= \mathbf{R}_z\left((i-1)\frac{\pi}{3}\right) \begin{bmatrix}
L_{j,i}\frac{\sin(q_{j,i})}{q_{j,i}}\\
0 \\
L_{j,i}\frac{1-\cos(q_{j,i})}{q_{j,i}} 
\end{bmatrix}, \quad  i=1, 3, 5.
\end{align}
 
To later model the soft arm's dynamics (see \Cref{sec:dynSoftArm}), we use a rigid model of the soft arm, known as Dynamically-Consistent Augmented Formulation~\cite{della2018dynamic, katzschmann2019dynamic, della2020model, toshimitsu2021sopra}. This model represents the soft arm with sufficient accuracy while remaining computationally efficient.
Therefore, we augment every single segment by four joints, consisting of two rotational and two translational joints. \Cref{fig:SoftArm} shows a single soft arm with the two segments and the model structure of the two augmented four degrees of freedom. The Denavit-Hartenberg (DH) parametrization of a single augmented segment can be found in \Cref{tab:DHparameters}. 

\begin{table}[htb]
\begin{center}
\setlength{\extrarowheight}{2pt}
\begin{tabular}{ c c c c c c  } 
 \toprule
 Link & a &$\alpha$ & d & $\upsilon$  \\ 
 \midrule
 1 & 0 & $\frac{\pi}{2}$ & 0 & $\frac{q_{j,i}}{2}$\\ 
 2 & 0 & $ 0 $ & $f(q_{j,i})$ & 0 \\
 3 & 0 & $-\frac{\pi}{2}$ &$f(q_{j,i})$ & $0$ \\ 
 4 & 0 & $0$ & 0 & $\frac{q_{j,i}}{2}$ \\
 \bottomrule
\end{tabular}
\caption{Denavit-Hartenberg (DH) parameters of a single augmented arm segment.}
\label{tab:DHparameters}
\end{center}
\end{table}

The state space representation $\xi_j\in \nR{8}$ of the augmented rigid arm is called the \textit{Augmented State} representation~\cite{della2018dynamic}. We connect the kinematic representation of the continuous curvature and the rigid robot through the continuously differential map~\cite{della2018dynamic}:
\begin{equation}
\vect{m}_i: \nR{2}\rightarrow\nR{8}
\end{equation}
with the segment mapping
\begin{equation}
\vect{m}_{j,i}(q_{j,i})=
\begin{bmatrix}
\frac{q_{j,i}}{2}\\
L_{j,i}\frac{\sin(\frac{q_{j,i}}{2})}{q_{j,i}}\\
L_{j,i}\frac{\sin(\frac{q_{j,i}}{2})}{q_{j,i}}\\
\frac{q_{j,i}}{2}
\end{bmatrix}.
\label{eq:augmentation}
\end{equation}
The mapping satisfies the condition that all endpoints of the continuous curvature segments of the soft arm coincide with their rigid body representation (see as well \Cref{fig:SoftArm}).

A single arm's configuration, consisting of two segments, is then connected by the map
\begin{equation}
\vect{m}_i(\vect{q}_i) = \left[\vect{m}_{1,i}^T\ \vect{m}_{2,i}^T\right]^T.
\label{eq:armConfiguration}
\end{equation}


\subsection{Dynamics}
\subsubsection{Dynamics of the rotors and rigid body}
To model the aerodynamics of a single spinning rotor, the standard in the literature (\emph{e.g.}, \cite{valavanis2008advances}) is followed. 
While rotating, the propeller generates a thrust force and a torque, applied in $O_{P_i}$ along $\vz_{P_i}$, that, expressed in $\bFrame$, are
\begin{align}
\vect{f}_i^B(f_i) &= f_i\vect{R}^B_{P_i} \vect{e}_3, \quad \text{for } i=1\dots6, \quad\text{and} \label{eq:propell_force} \\
\boldsymbol{\tau}_i^B(f_i) &= (-1)^{i-1} c_f^\tau f_i\vect{R}^B_{P_i} \vect{e}_3,	 
\quad \text{for } i=1\dots6 	 \label{eq:propell_drag} 
\end{align}
where $c_f^\tau>0$ is a constant parameter dependent on the type of propeller and $f_i$ is the intensity of the propeller force. $f_i$ is related to the spinning velocity by the relation
\begin{align}
 f_i = c_f w_i^2,
\end{align} 
where $c_f>0$ is a constant parameter relating the spinning velocity to the force.  $(-1)^i$ in \eqref{eq:propell_drag} represents 
the effect that neighboring propellers are designed to spin in opposite directions and thus generate opposite drags.

We can now compute the total force applied by the propellers to the aerial platform with respect to its center of mass, expressed in $\wFrame$
\begin{align}
\vect{f}^W(\vect{u_{1\dots6}}) = \vect{R}_B\sum_{i=1}^6\vect{f}_i^B(f_i) = \vect{R}_B\vect{F}_1 {\vect{u_{1\dots6}}},\label{eq:total_force}
\end{align}
where $ \vect{u_{1\dots6}}=[f_1\;f_2\;f_3\;f_4\;f_5\;f_6]^T$ and $\vect{F}_1\in\mathbb{R}^{3\times 6}$.

Summing all contributed moments (thrust contributions and drag moments), the moment applied to the platform with respect to $O_B$ and expressed in $\bFrame$ is
\begin{align}
{\boldsymbol \tau}^B(\vect{u_{1\dots6}}) =& \sum_{i=1}^6\left(\left(\vect{p}^B_{B,P_i}\times \vect{f}_i^B(f_i)\right) + {\boldsymbol\tau}_i^B(f_i) \right) \\
=&\ \vect{F}_2 \vect{u_{1\dots6}}.\label{eq:total_moment}
\end{align}
with $\vect{F}_2\in\mathbb{R}^{3\times 6}$.

\if
Using the Newton-Euler method, the equations of motion of the aerial platform due to the fixed propellers can be written as
\begin{align}
\begin{bmatrix}
m\ddot{\vect{p}}_B \\
\vect{J}\dot{{\boldsymbol \omega}}_B
\end{bmatrix}
= 
-\begin{bmatrix}
m g \vect{e}_3\\
{\boldsymbol \omega}_B \times \vect{J}{\boldsymbol \omega}_B
\end{bmatrix}
+
\begin{bmatrix}
\vect{f}^W\\
{\boldsymbol \tau}^B
\end{bmatrix}
\label{eq:newt-eul}
\end{align}
where $\vect{J} > 0$ is the $3\times 3$ inertia matrix of the rigid body with respect to $O_B$ and expressed in $\bFrame$, $m>0$ is the total mass of the platform, and $g>0$ is the gravitational acceleration.

Replacing~\eqref{eq:total_force} and~\eqref{eq:total_moment} in~\eqref{eq:newt-eul} 
we obtain 
\begin{align}
\begin{bmatrix}
m\ddot{\vect{p}}_B \\
\vect{J}\dot{{\boldsymbol \omega}}_B
\end{bmatrix}
=
-\begin{bmatrix}
m g \vect{e}_3\\
{\boldsymbol \omega}_B \times \vect{J}{\boldsymbol \omega}_B
\end{bmatrix}
+
\underbrace{
\begin{bmatrix}
\vect{R}_B\vect{F}_1(\boldsymbol\alpha)\\
\vect{F}_2(\boldsymbol\alpha)
\end{bmatrix}
}_{\vect{F}(\vect{R}_B,\boldsymbol\alpha)}
{\vect{u}_{246}}.
\end{align}

\fi
\subsubsection{Dynamics of the soft arm}\label{sec:dynSoftArm}
According to \cite{katzschmann2019dynamic,della2020model}, we can now present the dynamics of a single augmented rigid arm as
\begin{equation}
\vect{B}_{\xi_i}(\xi_i)\ddot{\xi_i}+\vect{C}_{\xi_i}(\xi_i,\dot{\xi}_i)\dot{\xi}_i+\vect{g}_{\xi_i}(\xi_i)=\vect{\tau}_{\xi_i}+\vect{J}_{\xi_i}^T \vect{f}_{\text{e}_i},
\label{eq:dynamicsArm}
\end{equation}
with $\xi_i$, $\dot{\xi}_i$, $\ddot{\xi}_i$ being the arm's configuration and its derivatives, $\vect{B}_{\xi_i}$ being the arm's inertial matrix, $\vect{C}_{\xi_i}$ the Coriolis and centrifugal terms and $\vect{g}_{\xi_i}$ the terms due to gravity. On the right side of the equation, we can find the applied inputs $\vect{\tau}_{\xi_i}$ and the effect of the external wrench $\vect{f}_{\text{e}_i}$ on the arm, through the end-effector Jacobian $\vect{J}_{\xi_i}$. To express \eqref{eq:dynamicsArm} with respect to the true inputs $q_{j,i}$ on the sub-manifold as defined in \eqref{eq:augmentation}, we can find the configuration derivatives  $\dot{\xi}_{i}$, $\ddot{\xi}_{i}$ w.r.t $q_{j,i}$ $\dot q_{j,i}$, $\ddot q_{j,i}$ as
\begin{align}
\dot{\xi}_{i}&=\vect{J}_{m_{j,i}}(q_{j,i})\dot q_{j,i}\\
\ddot{\xi}_{i}&=\dot{\vect{J}}_{m_{j,i}}(q_{j,i},\dot q_{j,i})\dot q_{j,i} + \vect{J}_{m_{j,}}(q_{j,i})\ddot q_{j,i}
\label{eq:jacobiansArm}
\end{align}
with $\vect{J}_{m_{j,i}}(q_{j,i})$ being the Jacobian of $m(q_{j,i})$ and being defined as $\vect{J}_{m_{j,i}}=\frac{\partial m_{j,i}}{\partial q_{j,i}}$, resulting in
\begin{equation}
\vect{J}_{m_{j,i}}(q_{j,i}) = \left[ \frac{1}{2} \ \ L_{c_{j,i}}(q_{j,i})\ \ L_{c_{j,i}}(q_{j,i})\ \ \frac{1}{2}\right]^T
\end{equation}

for \eqref{eq:armConfiguration} with $L_{c_{j,i}}(q_{j,i})\ = L_{j,i}\frac{q_{j,i}\cos\left(\frac{q_{j,i}}{2}\right)-2\sin\left(\frac{q_{j,i}}{2}\right)}{2q_{j,i}^2}$. Following \cite{katzschmann2019dynamic,della2020model}, the input constraints are projected through pre-multiplication with $J^T_{m_{j,i}}(q_{j,i})$ onto \eqref{eq:dynamicsArm} through \eqref{eq:jacobiansArm}, resulting in the compact form 
\begin{equation}
\vect{B}_{\text{arm}}(\vect{q})\ddot{\vect{q}}+\vect{C}_{\text{arm}}(\vect{q},\dot{\vect{q}})\dot{\vect{q}}+\vect{g}_{\text{arm}}(\vect{q})=\vect{\tau}+\vect{J}_{\text{arm}}^T(\vect{q}) \vect{f}_{\text{e}},
\end{equation}
with
\begin{equation}
\begin{cases}
\vect{B}_{\text{arm}}(\vect{q}) & = \vect{J}_m^T(\vect{q}) \vect{B}_\xi(m(\vect{q})) \vect{J}_m(\vect{q})\\
\vect{C}_{\text{arm}}(\vect{q},\dot{\vect{q}}) & = \vect{J}_m^T(\vect{q}) \vect{B}_\xi(m(\vect{q})) \dot{\vect{J}}_m(\vect{q})\\
&\ \ +\ \vect{J}_m^T(\vect{q}) \vect{C}_\xi\left(m(\vect{q}), \vect{J}_m(\vect{q})\dot{\vect{q}}\right) \vect{J}_m(\vect{q})\\
\vect{g}_{\text{arm}}(\vect{q}) & = \vect{J}_m^T (\vect{q})  \vect{G}_\xi(m(\vect{q}))\\
\vect{\tau} & = \vect{J}_m^T (\vect{q}) \tau_{\xi}\\
\vect{J}_{\text{arm}}(\vect{q})  & = \vect{J}_{\xi}(m(\vect{q})) \vect{J}_m(\vect{q})
\end{cases}
\end{equation}

where $\vect{f}_{\text{e}}$ is an external force and the $[]_{j,i}$ index is omitted for better readability.
 
\subsubsection{Dynamic Model of the Full Body}
 
We now compute the full body dynamics of our platform consisting of the three soft arms and the aerial main body. The dynamical model can be written in the compact Euler-Lagrange formulation:
 \begin{equation}
\vect{B}_{q}(\vect{q})\ddot{\vect{q}}+\vect{C}_{\vect{q}}(\vect{q},\dot{\vect{q}})\dot{\vect{q}}+\vect{g}_{\vect{q}}(\vect{q})=\vect{G}(\vect{q})\vect{u}+\vect{J}_e^T(\vect{q}) \vect{f}_{\text{e}},
\end{equation}
where $\vect{B}_q\in \nR{(6 + 2 \cdot 3) \times (6 + 2 \cdot 3)}$ is the positive definite inertia matrix, $\vect{C}_{q}(q,\dot q)$ maps the centrifugal and Coriolis terms, $\vect{g}_{q}(q)$ contains all gravity terms and $\vect{q}=[\vect{p}_B\ \vect{\eta}\ q_{1,1}\dots q_{2,5}]^T$ are the generalized coordinates of the system ($\eta$ is a minimal parametrization of $\vect{R}_B$, \emph{e.g.}, roll, pitch and yaw angles). $\vect{G}(\vect{q})$ maps the control inputs $\vect{u}=[\vect{u}_{1\dots6}\ \tau_1\ \tau_2\ \tau_3]^T$. $\vect{J}_e$ is the end-effector Jacobian, mapping the end-effector disturbances to the joint forces and torques. $\vect{B}_\vect{q}$ is constructed as
\begin{equation}
\vect{B}_\vect{q} = 
\begin{bmatrix}
\vect{B}_{B_{1,1}}& \vect{B}_{B_{1,2}} 	& \vect{B}_{1,3}		& \vect{B}_{1,4}		& \vect{B}_{1,5}  \\
\vect{B}_{B_{2,1}}& \vect{B}_{B_{2,2}}	& \vect{B}_{2,3}		& \vect{B}_{2,4}		& \vect{B}_{2,5}	      	\\
\vect{B}_{3,1}	& \vect{B}_{3,2}		& \vect{B}_{Arm 1}	& \vect{0}			& \vect{0}	      	\\
\vect{B}_{4,1}	& \vect{B}_{4,2}		& \vect{0}			& \vect{B}_{Arm 2}	& \vect{0}		 \\
\vect{B}_{5,1}	& \vect{B}_{5,2}		& \vect{0}			& \vect{0}			& \vect{B}_{Arm 3}
\end{bmatrix}
\end{equation}

with 
\begin{align*}
\vect{B}_{B_{1,1}} &= \left(m_B +\sum_{i=1}^3 m_{l_i}\right)\vect{I}_3\\
\vect{B}_{B_{2,2}} &= \vect{Q}^T\vect{H}_B\vect{Q} + \sum_{i=1}^3(m_{l_i}\vect{T}^T_B \vect{S}(\vect{R}_B p_{bl_i})^T \vect{S}(\vect{R}_B p_{bl_i}) \vect{T}_B \\
& +\vect{Q}^T \vect{R}_{l_i}^b\vect{H}_{l_i} \vect{R}_{p}^{l_i}\vect{Q})\\
\vect{B}_{B_{1,2}} &=\vect{B}_{B_{2,1}}^T=\sum_{i=1}^2(m_{l_i}\vect{S}(\vect{R}_B p_{bl_i})\vect{T}_B)\\
\vect{B}_{1,3..5} &=\vect{B}_{3..5,1}^T=\sum_{i=1}^2\left(m_{l_i}\vect{R}_B\vect{J}_p^{l_i}\right)\\
\vect{B}_{2,3..5} &=\vect{B}_{3..5,2}^T=\sum_{i=1}^2\left(\vect{Q}^T\vect{R}_{l_i}^b\vect{H}_{l_{i}}\vect{R}_B^{l_i}\vect{J}_O^{l_i}\right. \\
&\phantomrel{=} \hphantom{\vect{B}_{3..5,2}^T=} \left. -m_{l_i}\vect{T_B}^T\vect{S}(\vect{R}_B\vect{p}_{bl_i}^b)^T\vect{R}_B\vect{J}_P^{l_i}\right) \quad .
\end{align*}

For the lack of space we omit a detailed derivation of $\vect{B}_q$ and its components and refer the interested reader to \cite{lippiello2012cartesian}.

\section{Control}

In this first step, we aim at enabling a flying motion combined with a jumping locomotion, inspired by a friction-free bouncing ball. We therefore like to track a trajectory generated offline by a deformable bouncing ball. The goal is to track the ball's position and attitude with the drone's body center. 

\subsection{Main body state error}
We compute the state position error as 

\begin{equation}
\vect{e}_p = \vect{p}_d - \vect{p}_B
\label{eq:positionError}
\end{equation}

where $\vect{p}_d \in \nR{3}$ is desired position and 

\begin{equation}
\vect{e}_v = \vect{v}_d - \vect{v}_B
\label{eq:velocityError}
\end{equation}

is the translational velocity error with $\vect{v}_d \in \nR{3}$  being the desired velocity.

The rotational position error is computed as 

\begin{equation}
\vect{e}_R = \frac{1}{2}\left[\vect{R}_B^T \vect{R}_d - \vect{R}_d^T \vect{R}_B\right]_\vee
\label{eq:attitudeError}
\end{equation}

where $\vect{R}_d\in\nR{3\times3}$ is the desired orientation and $[]_\vee$ represents the inverse mapping from $SO(3)$ to $\nR{3}$. The rotational velocity error is

\begin{equation}
\vect{e}_\omega = \vect{R}_B^T\vect{R}_d\vect{\omega}_d - \vect{\omega}_B
\label{eq:attVelError}
\end{equation}

with $\vect{\omega}_d$ being the desired rotational velocity. Utilizing \cref{eq:positionError,eq:velocityError,eq:attitudeError,eq:attVelError}, we can now define 

\begin{align}
\vect{\nu}_p&= \vect{k}_p\vect{e}_p +\vect{k}_v \vect{e}_v + \vect{a}_d \label{eq:poserror}\\
\vect{\nu}_R&= \vect{k}_R\vect{e}_R +\vect{k}_\omega \vect{e}_\omega + \dot{\vect{\omega}}_d 
\label{eq:atterror}
\end{align}

 with $\vect{k}_{[]}\in\nR{3 \times 3}$ being positive diagonal definite proportional gain matrixes and $\dot{\vect{\omega}}_d,  \vect{a}_d  \in\nR{3}$ being the rotational and translational desired acceleration.
 
\subsection{Arm position control}
For the jumping motion we aim at keeping the arms at a constant position.

Given the kinematics in \eqref{eq:armKinematics}, we can compute the end-effector position in body frame from the position vector of the transformation matrix in \eqref{eq:armKinematics} as

\begin{equation}
\vect{p}_{\text{end}_i} = \mathbf{R}_z\left((i-1)\frac{\pi}{3}\right)[\vect{T}_{1,i}(q_{1,i})\vect{T}_{2,i}(q_{2,i})]^\ddagger
\label{eq:endeffpos}
\end{equation}

where $[]^\ddagger$ selects the position vector of the homogeneous transformation matrix. Next, we compute the state error of lateral arm positions as

\begin{equation}
\vect{\nu}_{arm_i} = k_{p_{arm}} (p_{d_{\text{end},i}}-p_{\text{end},i}) + k_{d_{arm}} (\dot{p}_{d_{\text{end}},i}-\dot{p}_{{\text{end},i}})
\label{eq:endefferror}
\end{equation}

where $p_{\text{end},i}$ is the second component of $\vect{p}_{\text{end}_i}$ and $p_{d_{\text{end},i}}$ is the desired end-effector position expressed in body frame and $k_{p_{arm}},\ k_{d_{arm}}$ are suitable gains.

\subsection{Closed loop controller}
We now propose the following control loop for tracking the desired body and end-effector trajectory

 \begin{equation}
\vect{u} = \vect{G}^\dagger(\vect{B}_{q}(\vect{q}){\vect{\nu}}+\vect{C}_{\vect{q}}(\vect{q},\dot{\vect{q}})\dot{\vect{q}}+\vect{g}_{\vect{q}}(\vect{q})),
\end{equation}

where $\vect{\nu} = [\nu_p^T\ \nu_R^T\ \nu_{arm_1}\ {\nu}_{arm_2}\ {\nu}_{arm_3}]^T$ are the terms from \eqref{eq:poserror}, \eqref{eq:atterror},  \eqref{eq:endefferror} and $[]^\dagger$ is the Moore-Penrose inverse.

\section{Experiments and Discussion}

We conducted an experimental set to demonstrate the capabilities of the system. First, we present the multimodal locomotion capabilities by tracking a combined jumping and flying trajectory that utilizes the full pose tracking and the advantages of storing energy in the arms during jumping motions. Second, we demonstrate the energy efficiency with respect to standard flying. We encourage the reader to appreciate the attached video for better visualization of both experiments.
To conduct the simulated experiments, we modeled the system using MATLAB-Simulink. The system dynamics are modeled by using Simscape\,\cite{miller2021simscape} including the flexible elements toolbox to emulate the soft arms and their dynamics. We chose a timestep of $t=\SI{0.002}{\second}$ for the simulation. To emulate contact forces, a rigid contact model has been used. 

\subsection{Experimental Results - Forward jumping trajectory}

\begin{figure}[t]
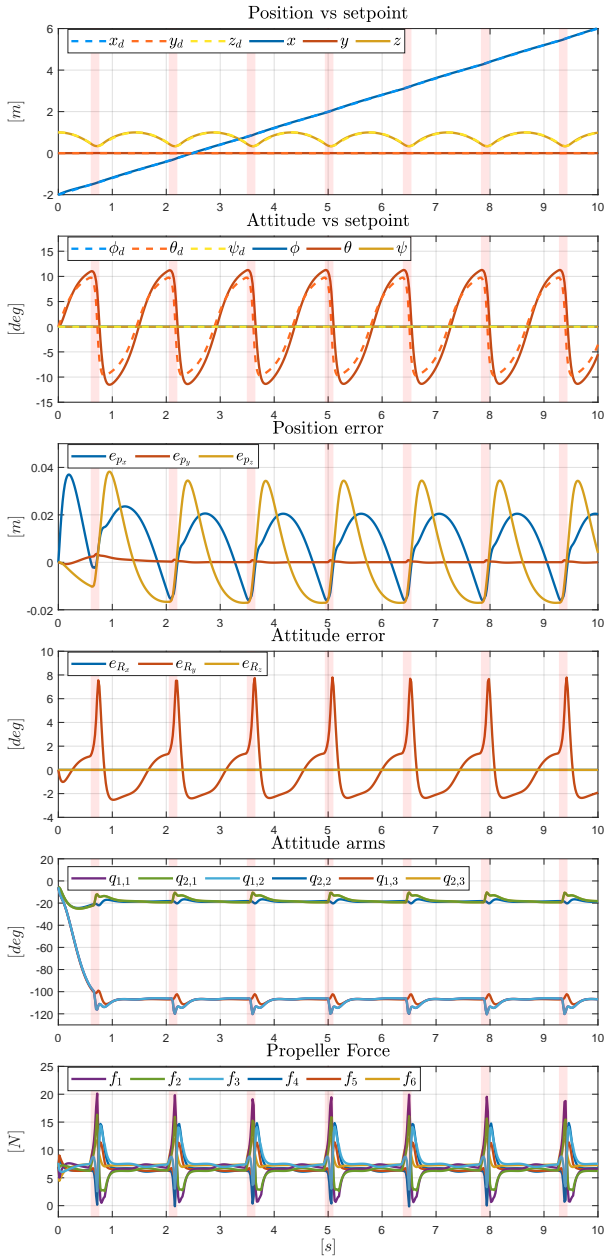

\centering
\figExpOnePos\\
\figExpOneAtt\\
\figExpOneeP\\
\figExpOneeR\\
\figExpOneArms\\
\figExpOneProps
\caption{Simulation results of a forward jumping trajectory. The contact times with the ground are highlighted in red. From top to bottom: a) Desired and actual body position, b) Desired and actual orientation, c) Position Error, d) Attitude error, e) Tilting angle of arms, f) Propeller forces.}
\label{fig:Exp1}
\vspace{-5mm}
\end{figure}

To conduct this experiment, we chose a predefined trajectory inspired by a forward jumping, friction-free deformable ball, but with a reduced gravitational acceleration ($g=30\% \cdot\SI{9.81}{\meter\per\square\second}$). The main body of the system tracks the position trajectory of the ball. 
The desired body orientation was selected as $\vect{R}_d=[\vect{x}_d\ \vect{y}_d\ \vect{z}_d]$ with $\vect{x}_d = \vect{v}_d/||\vect{v}_d||$, $\vect{y}=[0\ 1\ 0]^T$, $\vect{z}_d=\vect{x}_d \times \vect{y}$ and $\vect{y}_d=\vect{x}_d \times \vect{z}_d$.
The desired and the tracked position trajectory are plotted in \Cref{fig:Exp1}-first and the desired attitude and the tracked attitude are plotted in \Cref{fig:Exp1}-second. From the attitude plot, it becomes obvious how the main body rapidly changes its attitude when contacting the ground. The third and fourth plots represent the position and attitude error while tracking the trajectory. The controller can track the desired full trajectory very well, even in the cases of contact with a peak norm position error of $\norm{\vect{e}_p}\approx\SI{0.09}{\meter}$. This position error in ground contact is desired to store kinetic energy in the soft arms to accelerate later when jumping forward again. The peak norm attitude error is $\norm{\vect{e}_R}\approx\SI{8}{\degree}$, occurring during contact. The high attitude error along $e_{R_y}$ results from the contact forces of the end-effectors. Plot five of \Cref{fig:Exp1} shows the tilting angles of the arms. It is clear how the tilting angle is affected by the contact as the two soft arms in the back first touch the ground before the soft arm in front gets in contact. The last plot presents the propeller forces while tracking the trajectory.

\begin{figure}[t]
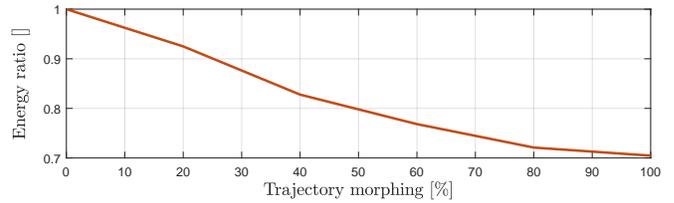

\centering
\figEnergy
\caption{We compare the energy consumption of the platform with respect to a morphing of the forward jumping trajectory. The morphing starts from a pure forward translation to a forward translation combined with a jumping maneuver. During the morphing, the vertical acceleration is linearly increased until 100$\% \cdot \SI{9.81}{\meter\per\square\second}$ are reached. The energy consumption is normalized with respect to the pure translation.}
\label{fig:HexaComparison}
\vspace{-5mm}
\end{figure}

To demonstrate the energy efficiency of the SMORS platform, we altered the gravitational acceleration $g$ of the presented trajectory from 0$\%$ to 100$\%$. This increase changes the trajectory from a pure lateral motion (0$\%$), over slow hopping to falling with gravitational acceleration (100$\%$). The interested reader is referred to the attached video. \Cref{fig:HexaComparison} depicts that with an increasing acceleration the energy consumption decreases continuously, requiring only $70\%$ energy with respect to continuous hovering. While this initial analysis does not take into account an increased mass of the SMORS platform compared to a conventional hexarotor, it demonstrates that the combined jumping-flying locomotion can preserve energy.

\section{Conclusion and Future Work}\label{SEC:Concl}
We presented the first partially soft multirotor system consisting of three rigid and three continuously soft arms. The soft arms contain aerodynamic and tendon actuators. In an experiment, we demonstrated the capabilities of such a system in multimodal locomotion task. 

In the next steps, we want to continue and extend the work by first building a physical prototype to test our design and control approach. Additionally, we will develop an optimal trajectory generation scheme to obey input saturation and minimize energy consumption over a multimodal locomotion trajectory.

\bibliographystyle{IEEEtran}
\bibliography{bibCustom}
\end{document}